\documentclass[runningheads,a4paper]{llncs}

\usepackage{amssymb}
\usepackage{graphicx}
\usepackage{url}

\begin{document}

\mainmatter  

\title{Survey on Vision-based Path Prediction}

\titlerunning{Survey on Vision-based Path Prediction}

\author{Tsubasa Hirakawa\inst{1}
\and
Takayoshi Yamashita\inst{1}
\and \\
Toru Tamaki\inst{2}
\and
Hironobu Fujiyoshi\inst{1}}
\authorrunning{Hirakawa et al.}
\tocauthor{T. Hirakawa, T. Yamashita, T. Toru, H. Fujiyoshi}

\institute{
Chubu University, Aichi 487-0027, Japan\\
\email{hirakawa@mprg.cs.chubu.ac.jp}\\
\email{yamashita@cs.chubu.ac.jp}\\
\email{hf@cs.chubu.ac.jp}
\and
Hiroshima University, Hiroshima 739-8527 Japan\\
\email{tamaki@hiroshima-u.ac.jp}
}
\maketitle

\begin{abstract}

Path prediction is a fundamental task for estimating how pedestrians or vehicles are going to move in a scene.
Because path prediction as a task of computer vision uses video as input, various information used for prediction, such as the environment surrounding the target and the internal state of the target, need to be estimated from the video in addition to predicting paths.
Many prediction approaches that include understanding the environment and the internal state have been proposed.
In this survey, we systematically summarize methods of path prediction that take video as input and
and extract features from the video.
Moreover, we introduce datasets used to evaluate path prediction methods quantitatively.

\end{abstract}


\section{Introduction}
\label{sec:intro}

Path prediction is the task of estimating the path, or trajectory,
along which a target (e.g., a pedestrian or vehicle) will move.
Predicting paths from video is an important task receiving much attention 
as it is expected to have many potential applications, such as
surveillance camera analysis, self-driving cars, and autonomous robot navigation.

Path prediction has to estimate much more information --- such as information of the surrounding environment, moving direction, and status of prediction targets --- than other simple image recognition tasks.
As a result, prediction methods are often built on top of other computer vision tasks, such as
pedestrian detection \cite{Weinland2011,Benenson2015},
pedestrian attribute recognition \cite{Deng2014},
and semantic segmentation \cite{Zhu2016}.
Moreover, in the prediction task, future observations of predicted paths are not available.
In tasks of pedestrian detection and tracking, observations from the past to the present are used to locate and track the target in the current frame of the video.
In contrast, the prediction task localizes and predicts the locations of the target in future frames of the video,
using observations made until the present time
and prior information on the surrounding environment and knowledge of the target motion.

Path prediction has been studied for decades in the field of robotics.
At stations and airports,
robots need to move without interfering with the many people present \cite{Ziebart2009}
and to plan a path of efficient motion in the environment.
Path prediction is necessary to achieve such tasks.
However, in addition to information from cameras, robots are able to use information from many types of sensor, such as a LIDAR sensor, to obtain the three-dimensional (3D) geometry of the scene.
The environment in which the robot can move around is sometimes explicitly given as an environment map.
The present survey is of path prediction methods involving video only as a computer vision task.

There is an alternative task called early recognition, which predicts future human behaviors in video.
This task predicts future actions in the video but is excluded from the survey because the predicted categories are discrete whereas predicted paths are sequences of continuous locations.

\begin{figure}[t]
\centering
\includegraphics[width=\linewidth]{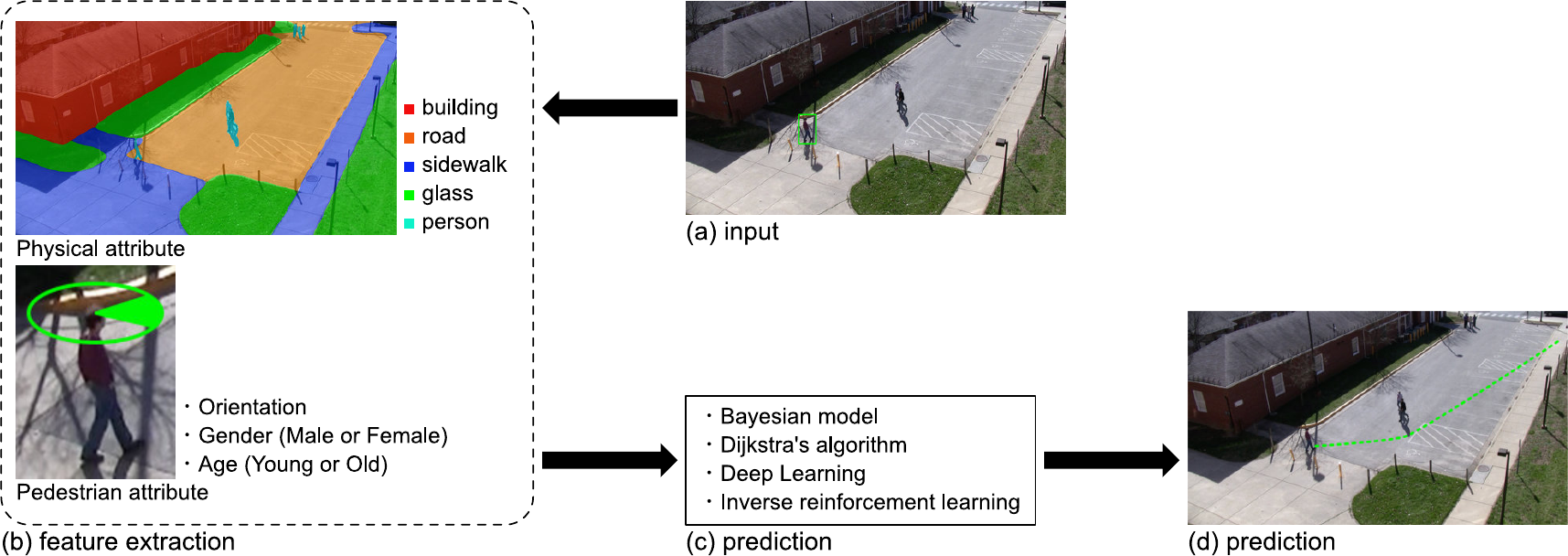}
\caption{
Overview of path prediction, modified from \cite{VIRATdataset}.
}
\label{fig:overview}
\end{figure}

As the task of path prediction in the field of computer vision is difficult and challenging,
a number of various methods have been proposed.
A common approach is shown in Fig \ref{fig:overview}.
As input, a video (or a frame of video) is given
in addition to the location of the target in the current frame or a sequence of locations over the past frames of several seconds.
Features useful for prediction are then extracted from the video (or frames) to predict the path in future frames.
There are two important parts to the overview of Fig. \ref{fig:overview}:
(b) feature extraction where many features are extracted to understand the environment and target;
and (c) path prediction where a variety of methods are proposed, categorized into four types.

In this paper, we survey path prediction methods taking video as input
and systematically summarize feature extraction and prediction approaches and datasets used for evaluation.
We explain feature extraction methods in section 2
and categorize prediction methods in section 3.
In section 4, we review datasets used in evaluating the performance of path prediction.
We conclude the survey in section 5.

\begin{table}[t]
\centering
\caption{
Categories of feature extraction for path prediction.
}
\label{tab:feature}

\begin{tabular}{c@{\hspace{2ex}}l@{\hspace{2ex}}l}
Feature & Types & Methods \\
\hline
Environment
     & Scene label   & 
     \parbox[c][4em][c]{5cm}{Stacked hierarchical labeling \cite{Munoz2010} \\
                             Superpixel-based MRF \cite{Yang2014} \\
                             Fully convolutional networks \cite{Long2015,Shelhamer2017}} \\ \cline{2-3}
     & Cost                 & 
     \parbox[c][3em][c]{5cm}{Bag of visual words \\
                             Spatial matching network \cite{Huang2016}} \\ \cline{2-3}
     & Global scene feature         &
     \parbox[c][3em][c]{5cm}{Pre-trained AlexNet \cite{Krizhevsky2012} \\
                             Siamese network \cite{Bromley1994}}\\
\hline
       & Location       & 
       \parbox[c][2em][c]{6cm}{HOG + SVM detector \cite{Dalal2005}} \\ \cline{2-3}
Target & Direction      & 
       \parbox[c][3em][c]{6cm}{Bayesian orientation estimation \cite{Enzweiler2010} \\ 
                               Orientation network \cite{Huang2016}} \\ \cline{2-3}
       & Attribute      & 
       \parbox[c][2em][c]{6cm}{AlexNet-based multi-task learning \cite{Wei2017}} \\ \cline{2-3}
       & Feature vector & 
       \parbox[c][2em][c]{6cm}{Mid-level patch features \cite{Singh2012}} \\
\hline
\end{tabular}

\end{table}

\section{Feature extraction from a video}
\label{sec:feature}

This section introduces methods of feature extraction from video for path prediction.
The path that the pedestrian takes is implicitly affected by many factors of the surrounding environment and the status of the pedestrian his self or herself.
The performance of path prediction is expected to be improve when using information that largely determines how the pedestrian decides the way to go.
Given the video, such information is extracted prior to the prediction.
Table \ref{tab:feature} presents information extracted from video for path prediction.
Such information can be broadly categorized into that of (1) the environment and (2) the target.

\subsection{Environmental features}
\label{sub:physical_attr}

The pedestrian decides the way and walks along a path while being affected by the surrounding environment.
For example, we usually walk along the sidewalk while avoiding obstacles on the way (e.g., parked cars and trash cans)
and drive a car on the roadway as is common social practice.
The movement of the target is dynamically affected by the environment, and environmental features are therefore extracted from the video.

\begin{figure}[t]
\centering
\includegraphics[width=\linewidth]{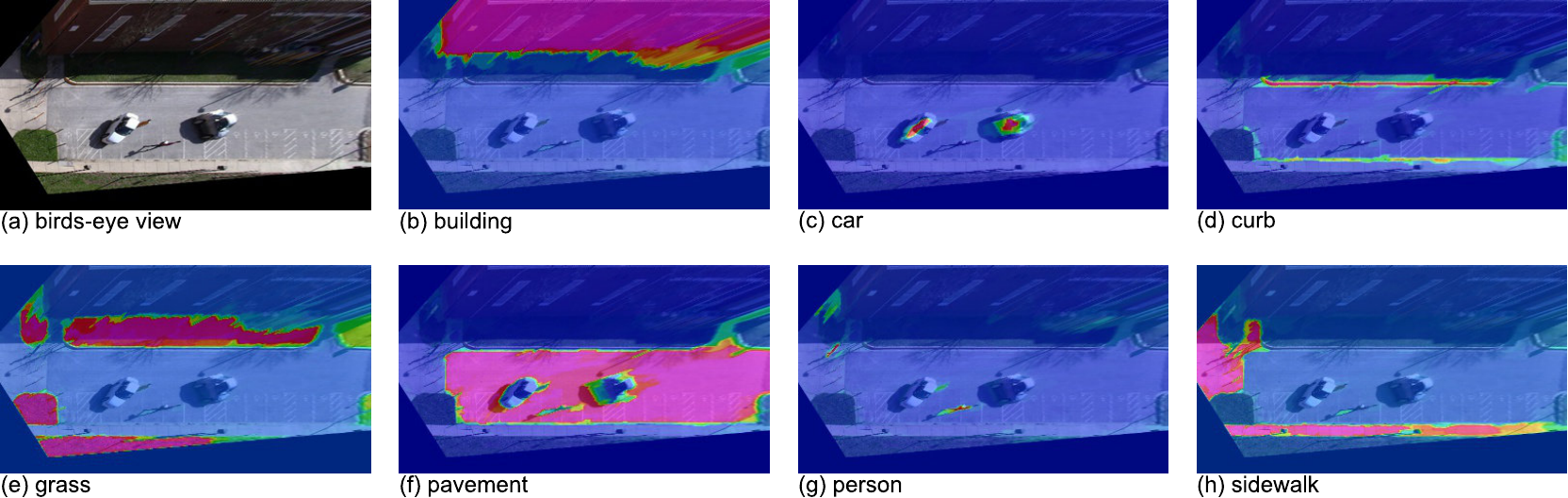}
\caption{
Examples of environmental attributes \cite{Kitani2012}
}
\label{fig:featuremap}
\end{figure}

Semantic segmentation \cite{Kitani2012,Ballan2016,Rehder2017,Lee2017} is a task of assigning an object class to each pixel, which is the most common task in understanding the environment in the field of computer vision.
Semantic segmentation can be conducted to estimate where obstacles exist in the scene and where there are regions available for walking.
Kitani et al. \cite{Kitani2012} assumed that pedestrian paths are mainly affected by the physical environment,
such as sidewalks, roadways, flower beds, and buildings, and
predicted posterior probabilities of each label using hierarchical segmentation \cite{Munoz2010} as shown in Figure \ref{fig:featuremap}.
These probabilities are used as feature vectors to form scene feature maps,
which are used for path prediction.
Rehder et al. \cite{Rehder2017} used segmentation results obtained using a fully convolutional network \cite{Long2015,Shelhamer2017} for prediction.

Alternative approaches do not explicitly use environmental features affecting paths
but implicitly represent probabilities of paths as cost (or reward) functions \cite{Walker2016,Huang2016}.
These methods create cost maps of the entire scene from cost functions independently estimated for each superpixel.
Walker et al. \cite{Walker2016} searched for patches that have similar texture from training samples using a nearest-neighbor approach,
and assigned the costs of the training samples to superpixels to generate cost maps of the scene.
Huang et al. \cite{Huang2016} proposed a convolutional neural network (CNN) called the spatial matching network, which estimates rewards of local regions by comparing similarity between the patch of the target and surrounding superpixel patches.

Yet another approach represents the scene as a single feature vector,
whereas the above approaches extract local features from superpixels.
Assuming that similar scenes prompt similar paths,
this approach retrieves similar scenes in a training dataset with feature vectors
to predict paths using the paths of the retrieved scenes.
To this end, CNNs are usually used to efficiently extract scene feature vectors because of the recent success of deep learning architectures.
In predicting paths in first-person video,
Park et al. \cite{Park2016} used AlexNet \cite{Krizhevsky2012} to extract features when retrieving scenes,
and transferred paths of the retrieved scenes for prediction.
Su et al. \cite{Su2017} used an AlexNet-based Siamese network \cite{Bromley1994} to retrieve features.

\subsection{Target features}
\label{sub:ped_attr}

\begin{figure}[t]
  \centering

    \parbox{.45\linewidth}{\centering\includegraphics[width=\linewidth]{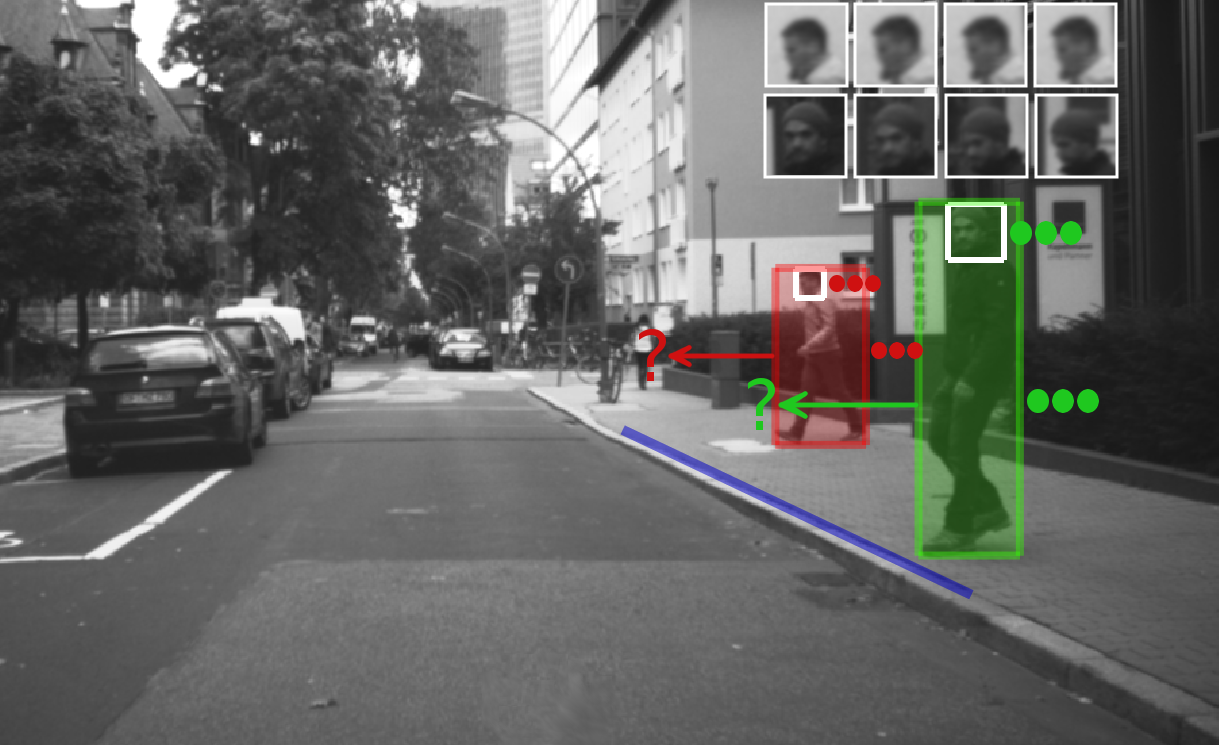}\\ (a)}
    \hspace{1pt}
    \parbox{.45\linewidth}{\centering\includegraphics[width=\linewidth]{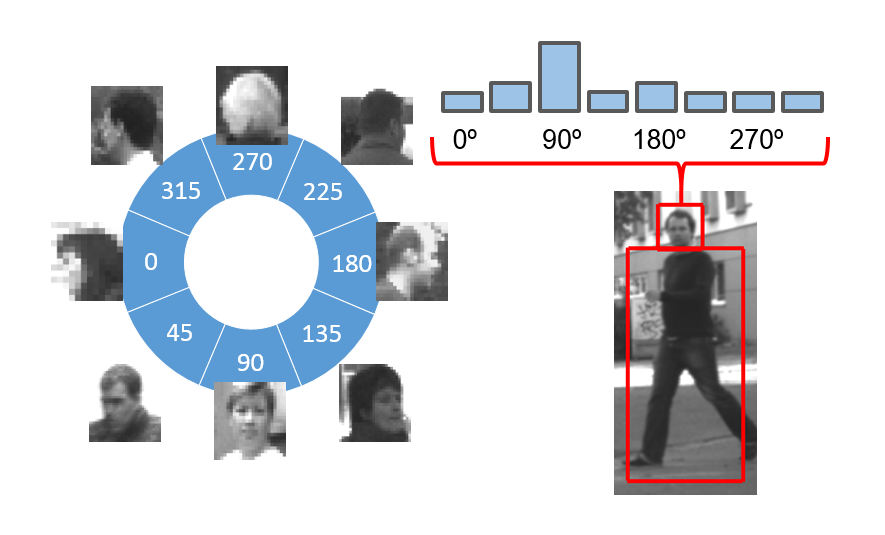}\\ (b)}

    \caption{
  Estimation of head orientation \cite{Kooij2014}. (a) Detection of heads and bodies of pedestrians. (b) Estimation of the orientation of the head in eight directions.}
  \label{fig:head_orientation}
\end{figure}

While environmental features strongly affect the target in terms of the path decision,
internal factors of the target are also important.
Specifically, attributes of the target, such as age, gender, and internal demand, affect the path decision.
We herein introduce methods for extracting target features.

The most common target feature is the orientation of the target \cite{Kooij2014,Huang2016,Wei2017}
because the estimated orientation can be used to predict in which direction the target is going.
In other words, the orientation constrains the moving direction of the target
and thus reduces errors of prediction.
Kooij et al. \cite{Kooij2014} detected pedestrians employing a histogram of oriented gradients (HOG) and support vector machine (SVM) \cite{Dalal2005}
and estimated the head orientation \cite{Enzweiler2010}
to predict the path of a pedestrian in front of a car on which a camera was mounted,
focusing on whether the pedestrian will stop before stepping forward onto the roadway as shown in Figure \ref{fig:head_orientation}.
If the head faces the camera, then the pedestrian is assumed to notice the car
and is predicted to slow down or stop before the roadway.

Physical attributes, such as age and gender, are also important to prediction.
When walking in places where there are a number of people, pedestrians take actions to avoid colliding with each other.
Aspects of such avoidance --- when and where pedestrians start to avoid others --- are different for pedestrians of different age and gender; e.g.,
a younger person walks faster and responds more rapidly to others than senior people.
Wei et al. \cite{Wei2017} used AlexNet to estimate the orientation, age, and gender of pedestrians as multi-task learning.
Estimated attributes are used in deciding the walking speed of pedestrians.

Walker et al. \cite{Walker2016} proposed unsupervised path prediction
by extracting mid-level feature vectors directly from patches containing the target,
instead of direct attributes.

\section{Prediction methods}
\label{sec:infer}

\begin{table}[t]
\centering
\caption{
Categories of path prediction methods
}
\label{tab:infer}

\begin{tabular}{ccccccccc}
     &      &      &      &          &      &      & \multicolumn{2}{c}{feature} \\
Category & Paper & Year & Method & Scene & Input & Output & Env. & Target \\ \hline
Bayesian     & 
               \cite{Schneider2013} & 2013 & KF & Car & Coord. & Coord. &  &   \\
             & 
               \cite{Kooij2014} & 2014 & DBN & Car & Video & Coord. & \checkmark & \checkmark  \\
             & 
               \cite{Ballan2016} & 2016 & DBN & Top view & Video & Coord. & \checkmark &   \\ \hline
Energy       & 
               \cite{Xie2013} & 2013 & Dijkstra & Top view & Video & Distribution  & \checkmark &   \\
Minimization & 
                \cite{Walker2016} & 2014 & Dijkstra & Surveillance & Video & Distribution & \checkmark & \\
             & 
                \cite{Huang2016} & 2016 & Dijkstra & Surveillance & Image & Distribution & \checkmark & \checkmark  \\ \hline
DL           & 
               \cite{Yi2016} & 2016 & CNN & Surveillance & Coord. & Coord. &  &   \\
             & 
               \cite{Alahi2016} & 2016 & LSTM & Top view & Coord. & Coord. &  &  \\
             & 
               \cite{Fernando2017a} & 2017 & LSTM & Top view & Coord. & Coord. &  & \\
             & 
               \cite{Fernando2017b} & 2017 & LSTM & Top view & Coord. & Coord. &  & \\
             & 
               \cite{Lee2017} & 2017 & RNN Enc.-Dec. & Car & Video & Coord. & \checkmark & \\ \hline
IRL          & 
                \cite{Kitani2012} & 2012 & IRL & Top view & Video & Distribution & \checkmark & \\
             & 
               \cite{Lee2016} & 2016 & IRL & Top view & Video & Distribution & \checkmark & \\
             & 
               \cite{Bokhari2017} & 2016 & IRL & First person & Video & Distribution &  & \checkmark \\
             & 
               \cite{Rhinehart2017} & 2017 & IRL & First person & Video & Distribution &  & \checkmark \\
             & 
                \cite{Wei2017} & 2017 & IRL & Surveillance & Image & Distribution &  & \checkmark  \\
             & 
               \cite{Rehder2017} & 2017 & IRL & Car & Video & Distribution & \checkmark &  \\ \hline
Others       & 
               \cite{Keller2014} & 2014 & Optical flow & Car & Video & Coord. &  &  \\
             & 
               \cite{Rehder2015} & 2015 & Markov process & Car & Video & Coord. & \checkmark &   \\
             & 
               \cite{Park2016} & 2016 & Data driven & First person & Video & Coord. & \checkmark &  \\
             & 
                \cite{Su2017} & 2017 & Data driven & First person & Video & Coord. & \checkmark &  \\
             & 
               \cite{Yamaguchi2011} & 2011 & Social force & Top view & Video & Coord. &  &  \\
             & 
              \cite{Robicquet2016} & 2016 & Social force & Top view & Video & Coord. &  &  \\
 \hline
\end{tabular}

\end{table}

Path prediction follows feature extraction from video.
Table \ref{tab:infer} summarizes methods of prediction,
categorized according to their approach.
This section describes each category and its properties.

\subsection{Bayesian models}
\label{sub:prob_model}

\begin{figure}[t]
  \centering
  \includegraphics[width=0.5\linewidth]{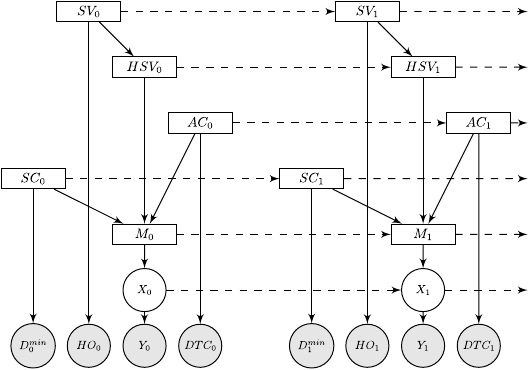}
  \caption{
  Graphical model of a DBN with an SLDS \cite{Kooij2014}
  }
  \label{fig:dbn}
\end{figure}

The first approach uses online Bayes filters, such as Kalman filters (KFs) and particle filters,
and infers the model to predict paths.
Such modeling introduces internal states and observations as variables,
and defines probabilistic models by assuming that the observations are the internal states contaminated by noise.
This approach iterates the prediction step that computes the current internal states from the previous states,
and the update step that updates the current states with the observations.
In a common setup,
internals states are actual coordinates of pedestrians,
and observations are coordinates obtained by pedestrian detection.
This is person tracking if we apply the approach to track from the past to present,
and path prediction if we only repeat the prediction step to obtain the sequence of coordinates
of the pedestrian, without the update step; i.e., there are no future observations.

Schneider et al. \cite{Schneider2013} used the extended KF to update the internal state
of the pedestrian in front of a car.
This was an early work of path prediction and showed what kind of primitive information (e.g., the walking speed and acceleration) is useful for path prediction.

Instead of using online Bayes filters,
some works have used a dynamic Bayesian network (DBN) \cite{Kooij2014,Ballan2016}.
Kooij et al. \cite{Kooij2014} considered a more restricted case;
estimating if the pedestrian will walk across a roadway 
in front of a car on which a camera is mounted.
They defined
a DBN model with a switching linear dynamical system (SLDS)
that is shown in Figure \ref{fig:dbn}
and that uses features extracted from the movie, such as
the pedestrian's head orientation, distance to the car, and distance between the pedestrian and roadway.
This method performs better than using coordinates of pedestrian detection only.

\subsection{Energy minimization}
\label{sub:energy}

The Bayes approach described above is on-line in which it estimates the coordinates of the pedestrian frame by frame in the video.
Another (off-line or batch) approach is an energy minimization approach that estimates the entire sequence of coordinates at the same time.
This approach constructs a two-dimensional grid graph of the scene and assigns costs for moving to edges in the graph,
and then finds the combination of edges that gives the minimum energy.
This is formulated as a shortest path problem solved employing the Dijkstra method.
The prediction accuracy is therefore largely affected by how the cost is defined.

Huang et al. \cite{Huang2016} proposed a path prediction method using a single image.
First, a patch containing the target is extracted to estimate the orientation of the target.
Next, the cost for moving across the location of the patch is estimated by
comparing the texture of surrounding patches.
In addition to this cost, the estimated orientation of the target is used as a constraint
and added to the edge weights.
Walker et al. \cite{Walker2016} compared the texture of superpixels
using patches along the path that the target traced
without involving any training procedure.

Appearance information (texture) of the scene can be used to define the cost function,
but objects in the scene can also be used.
Xie et al. \cite{Xie2013} assumed that pedestrians have decided their goal (e.g., a food trunk)
according to their potential demands (hunger), and defined cost maps where
the pedestrians are attracted to objects in the scene.

\subsection{Deep learning}
\label{sub:deep_model}

Deep learning methods such as those involving the CNN and long short-term memory (LSTM)
have been used for path prediction since the emergence of deep learning frameworks.
Methods of this type take as input the series of coordinates of the target over the last several frames,
and produce a series of target coordinates in several successive frames.
Feature extraction, described in the last section, is not explicitly performed
as feature extraction and prediction are not explicitly separated in deep learning models.

Several methods thus use LSTM to deal with paths,
which are sequences of two-dimensional coordinates, have been proposed.
Alahi et al. \cite{Alahi2016} proposed
the social-pooling (S-pooling) layer for avoiding collisions between pedestrians.
A pedestrian is represented by LSTM,
and hidden layer outputs of LSTMs of other people are connected to the S-pooling layer of the pedestrian.
This layer allows the LSTM of the pedestrian to represent the spatial relationship with nearby people (e.g., the distance to each other),
and thus predict the path avoiding collision.

LSTM has a limitation of long-term memory; i.e.,
paths in the distant future are difficult to predict.
Fernando et al. \cite{Fernando2017b} assumed the necessity of more elaborate long-term memory,
and proposed the tree memory network that hierarchically selects useful information of the past stored in memory cells and
performs better than other LSTM models.

Besides LSTM, the CNN is also used to directly make predictions.
Yi et al. \cite{Yi2016} proposed the behavior-CNN that predicts the future path from the past path.
This method first creates three-dimensional sparse data
whose channels store the pedestrian two-dimensional coordinates of the last several frames.
The sparse 3D data are encoded using convolution and pooling layers
and then decoded using deconvolution layers.
They also added location bias maps to each channel of encoded information
to account for different behaviors at different locations in the scene,
such as the locations of entrances and obstacles.

\subsection{Inverse reinforcement learning}
\label{sub:irl_model}

\begin{figure}[t]
    \centering
    \includegraphics[width=\linewidth]{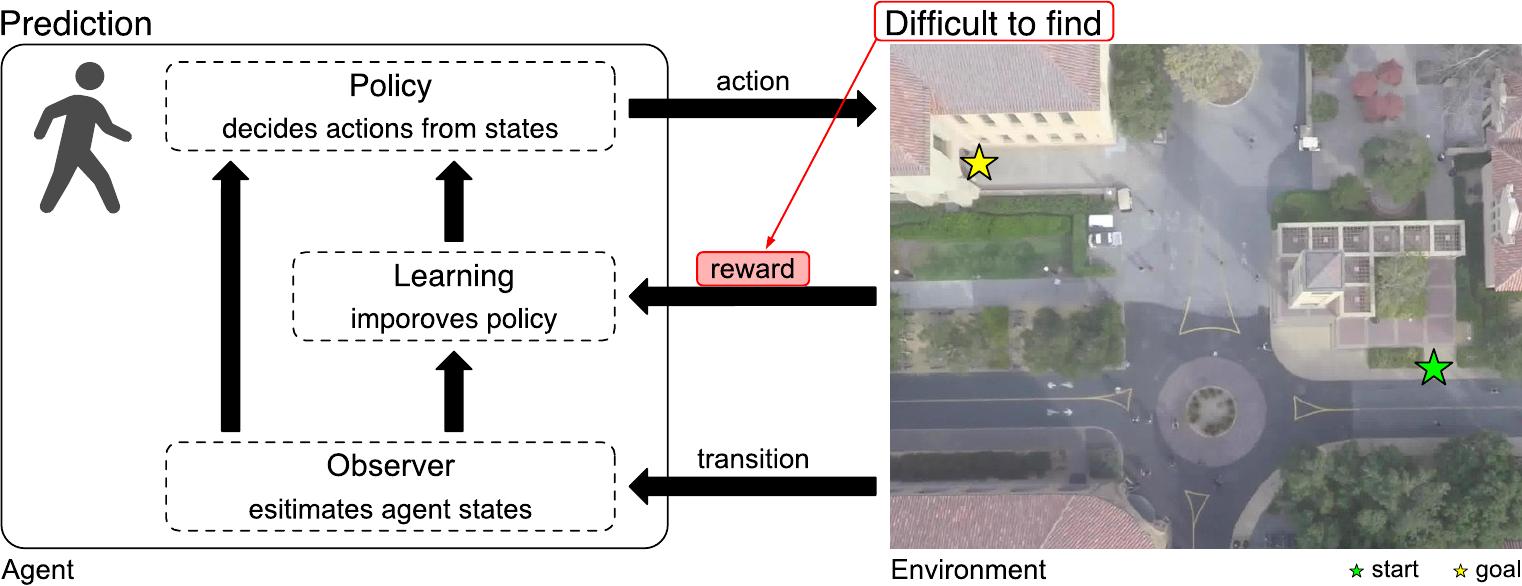}
    \caption{
    Overview of RL, modified from \cite{Robicquet2016}.
    }
    \label{fig:rl_overview}
\end{figure}

The three approaches above are examples of supervised or unsupervised learning,
while the approach presented here is an example of reinforcement learning (RL).
RL learns a policy to decide actions to be taken by an agent
under the current status in an environment.
RL is usually defined as a Markov decision process
that learns the optimal policy
to allow the agent to take the best actions maximizing the reward.
Figure \ref{fig:rl_overview} shows that
an agent of RL is the target of prediction,
an environment is the scene given as video,
a status is the pedestrian location,
and an action is the movement of the pedestrian.

RL needs to define the reward of the action of moving from one state to another,
which indicates how good the action taken by the agent is.
However, it is difficult to explicitly define the reward function
for practical problems such as the path prediction task.
This problem is called the reward design problem, and inverse reinforcement learning (IRL) is one approach taken to solve the problem.
IRL estimates rewards that reproduce optimal sequences of actions,
and decides actions of the agent in the test phase with the estimated reward
so that the agent can take similar actions.

IRL has been used to learn and control the optimal motion of robots \cite{Ziebart2009}.
Kitani et al. \cite{Kitani2012} first introduced IRL to vision-based path prediction.
Instead of estimating target locations,
they estimated actions that the agent may take at a certain time or location,
and predicted possible paths by sequentially applying the estimated actions to the current target location.
This task is therefore called activity forecasting, in contrast to path prediction that directly estimates locations of the target in the future.
Activity forecasting is a much more complex and challenging task than path prediction
while it has great potential in terms of having a variety of predictions adapted to each possible application.

Kitani et al. \cite{Kitani2012} assumed that the physical attributes of a scene strongly affect pedestrian paths,
and used scene attributes estimated by semantic segmentation as feature maps.
Rewards of each scene attribute are defined by the inner product of the feature maps and weight vectors,
and the optimal weights are estimated from training data.
For prediction, a sequence of actions that arrive at the predefined goal is generated
by giving the goal and the current location of the target pedestrian.
Lee et al. \cite{Lee2016} used a similar approach to predict paths of football players in a game video.
Wei et al. \cite{Wei2017} introduced a game theory called fictitious play
to predict paths of multiple pedestrians who arrive at a goal
while avoiding collisions between pedestrians.

Without any predefined goals,
Rehder et al. \cite{Rehder2017} proposed the destination network to estimate the goal of the target
using the last several frames. The estimated goal and the environmental attributes obtained using a fully convolutional network 
were used to predict pedestrian paths.

For first-person vision,
Bokhari et al. \cite{Bokhari2017} used objects held by a person and the object states
to predict goals in the future. While this work considered a limited scene (e.g., a kitchen),
Rhinehart et al. \cite{Rhinehart2017} dealt with wider areas, such as a home including a kitchen, bathroom, and living room.

\subsection{Other approaches}
\label{sub:other}

\begin{figure}[t]
  \includegraphics[width=\linewidth]{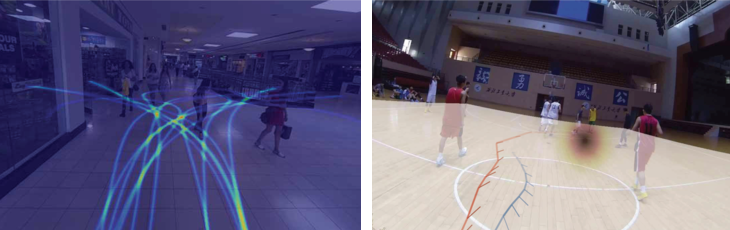}
  \caption{
  Prediction from first-person videos; (left) \cite{Park2016}, (right) \cite{Su2017}.
  }
  \label{fig:fps_prediction}
\end{figure}

Most prediction methods can be categorized into one of the four approaches described above,
but there are other approaches.

The social force model \cite{Helbing1995} assumes energy called a ``social force'' that acts between pedestrians
and objects in the scene, and generates pedestrian movement through interaction via the force.
Yamaguchi et al. \cite{Yamaguchi2011} proposed a model with additional states,
such as the preferences of pedestrians, walking speeds, goals, and the existence of other people walking together.
This work was motivated by a desire to improve the accuracy of pedestrian tracking,
but performed path prediction to evaluate the proposed model.
Robicquet et al. \cite{Robicquet2016} proposed social forces of multiple classes for avoiding collisions.
They estimated ``social sensitivity features'' using the distances between other people,
and applied K-means clustering of the features to get several clusters of avoidance behaviors.
The cluster of the target behavior of avoidance was estimated using the target feature,
and paths of the cluster were then projected back to the scene for prediction.

Optical flow extracted from car-mounted cameras was used by
Keller et al. \cite{Keller2014} to predict pedestrian paths.
They used optical flows over the last several frames and computed orientation histograms as motion features of pedestrians.
The sequence of histograms was used to retrieve similar scenes in the training set,
and paths of the retrieved scenes were then mapped back to the scene for prediction.

The use of the Markov process framework was proposed by Rehder et al. \cite{Rehder2015}.
They used normal and von-Mises distributions to represent the state (location) and speed of the pedestrian,
and sequentially estimated the state by taking products of these distributions at each time step for prediction.
To improve accuracy, the goal of the pedestrian was estimated from environmental attributes
to constrain the direction of motion.

The retrieval-based approach shown in Figure \ref{fig:fps_prediction} was proposed by Park et al. \cite{Park2016}
to predict the future path in a video showing the first-person view.
They first extracted scene features using AlexNet
and then found similar scenes in the training set by comparing extracted features.
Paths of retrieved training samples were mapped onto the video.
They predicted paths even in scenes with occlusions by estimating regions behind occluding objects, such as walls and obstacles.
Su et al. \cite{Su2017} extended this work to the prediction of multiple basketball players in a game scene.
In one first-person video,
they estimated the region of ``joint attention'' to which multiple players commonly paid attention.
Multiple paths were predicted by selecting the optimal path of each player
and by minimizing an objective function defined by the estimated joint attention region, locations of players,
and paths projected back to the scene.

\begin{table}[p]
\centering
\caption{
Comparison of datasets
}
\label{tab:dataset_stat}
\scriptsize

\begin{tabular}{l@{ }ccccccc}
& Year & URL & {\#}People & Viewpoint & {\#}Scenes & Oher targets 
& \parbox[b][][b]{1.5cm}{Additional\\ information} \\ \hline
\parbox[c][2em][c]{2cm}{UCY \cite{UCYdataset}} & 2007 & 1 & 786 & Top view & 3 & -- & -- \\ \hline
\parbox[c][2em][c]{2cm}{ETH \cite{ETHdataset}} & 2009 & 2 & 750 & Top view & 2 & -- & -- \\ \hline
\parbox[c][4em][c]{2cm}{Edinburagh Informatics Forum \cite{EdinburghDataset}} & 2009 & 3 & 95,998 & Top view & 1 & --  & -- \\ \hline
\parbox[c][5em][c]{2cm}{Stanford\\ Drone \cite{Robicquet2016}} & 2016 & 4 & 11,216 & Top view & 8 
& \parbox{1.7cm}{Bikers,\\ skateboarders,\\ cars, buses,\\ golf carts} & -- \\ \hline
VIRAT \cite{VIRATdataset} & 2011 & 5 & 4021 & surveillance & 11 & Car, bike 
& \parbox[c][4em][c]{2cm}{Object coordinates, Activity category} \\ \hline
\parbox[c][3em][c]{2cm}{Town Centre\\ \cite{TownCentreDataset}} & 2011 & 6 & 230 & Surveillance & 1 & -- 
& Head coordinates \\ \hline
\parbox[c][3em][c]{2cm}{Grand Central\\ Station \cite{GrandCentral}} & 2015 & 7 & 12,600 & Surveillance & 1 & -- & -- \\ \hline
\parbox[c][2em][c]{2cm}{Daimler \cite{Schneider2013}} & 2013 & 8 & 68 & Car & -- & -- 
& Stereo camera \\ \hline
KITTI \cite{KITTI} & 2012 & 9 & 6336 & Car & -- & Car 
& \parbox[c][3em][c]{2cm}{Stereo camera, LIDAR, Map} \\ \hline
\parbox[c][2em][c]{2cm}{EgoMotion \cite{Park2016}} & 2016 & -- & -- & First person & 26 & -- & Stereo \\ \hline
\parbox[c][4em][c]{2cm}{First-person Continuous Activity \cite{Rhinehart2017}} & 2017 & -- & -- & First person & 17 & -- & Object information \\ \hline
\end{tabular}

{\scriptsize
\begin{flushleft}
1: {\url{https://graphics.cs.ucy.ac.cy/research/downloads/crowd-data}}\\
2: {\url{http://www.vision.ee.ethz.ch/en/datasets/}}\\
3: {\url{http://homepages.inf.ed.ac.uk/rbf/FORUMTRACKING/}}\\
4: {\url{http://cvgl.stanford.edu/projects/uav\_data/}}\\
5: {\url{http://www.viratdata.org/}}\\
6: {\url{http://www.robots.ox.ac.uk/~lav/Papers/benfold_reid_cvpr2011/benfold_reid_cvpr2011.html}}\\
7: {\url{http://www.ee.cuhk.edu.hk/~xgwang/grandcentral.html}}\\
8: {\url{http://www.gavrila.net/Datasets/Daimler_Pedestrian_Benchmark_D/daimler_pedestrian_benchmark_d.html}}\\
9: {\url{http://www.cvlibs.net/datasets/kitti/}}\\
\end{flushleft}
}

\end{table}

\section{Datasets}
\label{sec:dataset}

\begin{figure}[t]
\centering
\includegraphics[width=\linewidth]{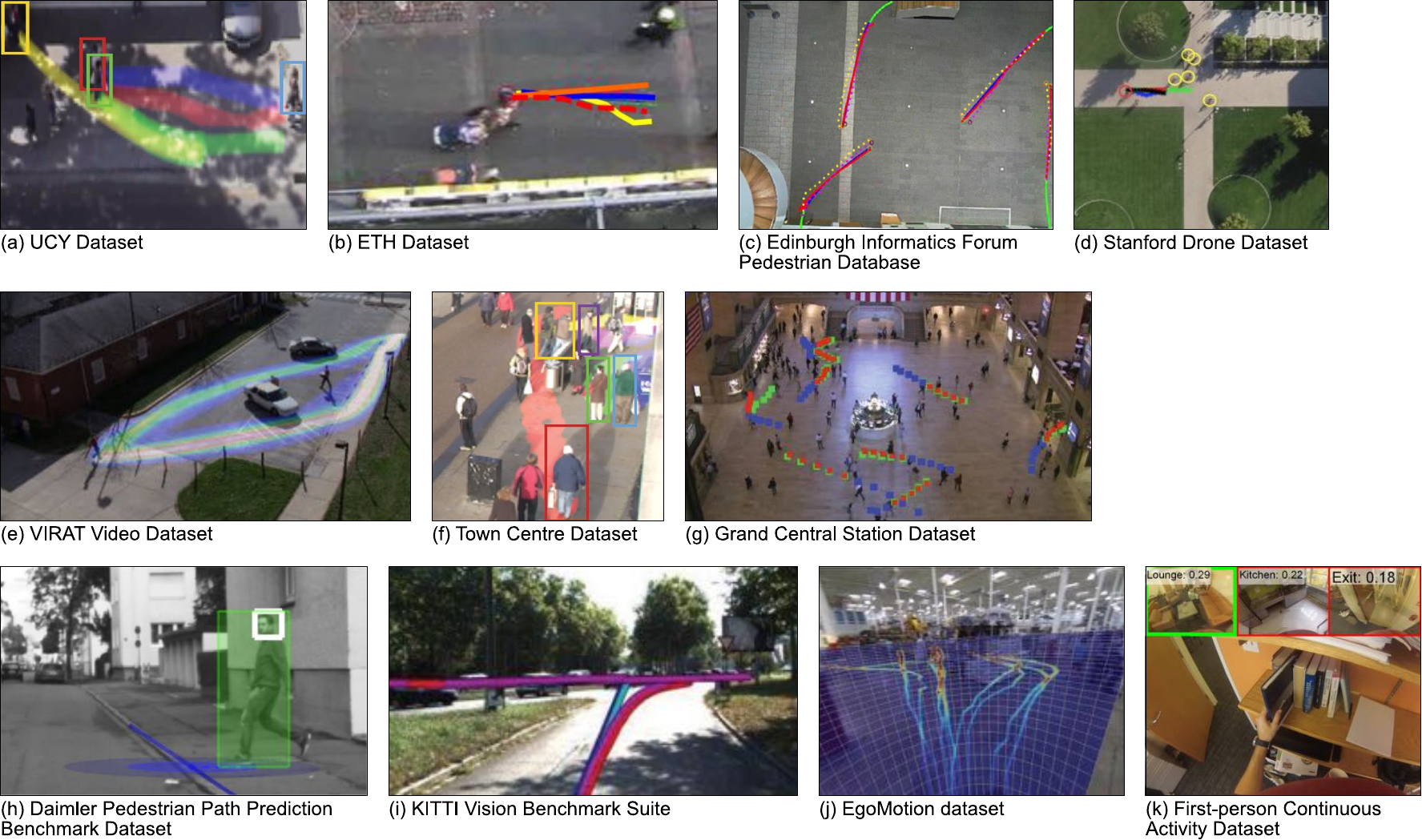}
\caption{
Datasets and results of prediction, taken and modified from \cite{Wei2017,Alahi2016,Fernando2017b,Robicquet2016,Kitani2012,Yi2016,Kooij2014,Lee2017,Park2016,Rhinehart2017}.
}
\label{fig:dataset}
\end{figure}

This section briefly introduces datasets used to evaluate path prediction methods.
Various datasets have been used as shown in Table \ref{tab:dataset_stat} and Figure \ref{fig:dataset}.
The diversity of datasets is due to the difficulty of using a single universal dataset
for many different conditions, e.g., different numbers of scenes 
and paths needed for learning and different types of scenes.
We therefore categorize datasets into four categories in terms of the viewpoint of the camera.

\subsection{Videos of entire scenes}

The most commonly used type of dataset is video that captures the entire scene
taken by a wide-angle camera (for surveillance) at stations and market places.
These datasets are usually used to evaluate pedestrian tracking methods;
however, they are also used in evaluating path prediction because
sequences of pedestrian locations are given as the ground truth.

\subsection*{Top view}

The UCY Dataset \cite{UCYdataset} and ETH Dataset \cite{ETHdataset} contain videos of pedestrians walking along streets
where no other moving objects exist, which is a relatively simple situation compared with situations of other datasets.
The Edinburgh Informatics Forum Pedestrian Database \cite{EdinburghDataset} consists of videos
of pedestrians walking at the campus of the University of Edinburgh taken by a fixed camera.
This dataset is large and has more than 90,000 paths.

The above datasets are constructed for pedestrian tracking and crowd behavior analysis,
while the Stanford Drone Dataset \cite{Robicquet2016} focuses on path prediction.
This dataset has videos taken by drones flying at eight sites of Stanford University,
and provides annotations of moving objects, such as cyclists, skateboarders, and cars, as well as pedestrians.

\subsection*{Surveillance}

Videos in the datasets described above are taken from a top view,
while videos in the datasets shown in Figure \ref{fig:dataset}(e, f) are taken from a bird's eye view;
i.e., the videos are taken by surveillance cameras looking downward at an angle.
The physical attributes of pedestrians are observable in these videos and can be used for prediction.
The VIRAT Video Dataset \cite{VIRATdataset} contains videos taken by surveillance cameras at parking lots,
and provides the locations of pedestrians, cars, and objects in the scene
and labels of activities, such as getting into a car and opening a trunk.
It contains 11 scenes, which is the largest number of scenes 
among the datasets of surveillance cameras in Table \ref{tab:dataset_stat}.
The Town Centre Dataset \cite{TownCentreDataset} contains videos of pedestrians
and provides bounding boxes of each pedestrian as well as labels of head locations of pedestrians.

The Grand Central Station Dataset \cite{GrandCentral} contains videos
taken by a fixed camera mounted at a station, as shown in Figure \ref{fig:dataset}(g).
It has a single scene but is complex owing to the many people appearing in and disappearing from the scene
because the motivation is to analyze the behaviors of many pedestrians.

\subsection{Car-mounted cameras}

Datasets of videos taken by cameras mounted on vehicles are used because path prediction is studied
with the aim to develop automated driving.
In this case, cameras are mounted in the front of the car to look forward,
and the main objective is to predict paths of pedestrians in front of the car.

The Daimler Pedestrian Path Prediction Benchmark Dataset \cite{Schneider2013}
consists of videos taken by car-mounted cameras.
There are four classes of cases, including cases that the pedestrian walks across the roadway and cases that the pedestrian stops walking to avoid an accident.
In addition to the videos themselves, depth information is available as the videos are taken by stereo cameras.
There are relatively few pedestrians; however, the dataset contains videos that are rare in other datasets,
such as videos of pedestrians crossing in front of moving cars.

The KITTI Vision Benchmark Suite \cite{KITTI} was constructed for the Intelligent Transport System,
and is used for various evaluations such as those of the detection of pedestrians, vehicles, and white lines on the road.
It contains not only RGB images but also stereo images, LIDER 3D data, GPS locations, and street maps,
and it is therefore useful for path prediction that uses rich information to understand the environment.

\subsection{First-person view}

Unlike videos of entire scenes and taken by car-mounted cameras for predicting paths of targets in the scene,
videos taken from the first-person view are used to predict the path of the person taking the video.
Park et al. \cite{Park2016} used first-person videos taken by wearable cameras moving through indoor and outdoor
environments of 26 different scenes, such as on a street and inside a store.
Rhinehart et al. \cite{Rhinehart2017} collected first-person videos taken by a person walking around office environments
and assumed that an object held by the person (e.g., a mug or towel) indicate where the person is going (e.g., the kitchen or bathroom).

\section{Conclusions}
\label{sec:conclusion}

We reviewed vision-based path prediction methods and common datasets.
We first categorized feature extraction methods of features used for prediction attributed to the environment or target appearance and dynamics.
We then grouped prediction methods according to the approach taken.
Bayesian methods define probabilistic models of the path and sequentially estimate internal states.
Energy minimization methods define a two-dimensional grid graph by computing possibilities of pedestrians to move in each local region, and then solve the shortest-path problem.
Deep learning methods take a series of locations of the target over the past several seconds and output a series of future locations.
IRL uses the policy and reward estimated from training samples and then selects actions iteratively to produce a future path.
These approaches are of course not exclusive and often used in combination \cite{Lee2017}.
Finally, we summarized datasets used in evaluating prediction methods.
Some datasets are used for pedestrian detection and tracking, while others are used for path prediction.

\section*{Acknowledgments}
This work was supported in part by JSPS KAKENHI under grant number JP16H06540.

\bibliographystyle{splncs}
\bibliography{reference/ref,reference/dataset,reference/robotics,reference/basic_task,reference/english}

\begin{thebibliography}{10}

\bibitem{Weinland2011}
Weinland, D., Ronfard, R., Boyer, E.:
\newblock A survey of vision-based methods for action representation,
  segmentation and recognition.
\newblock Computer Vision and Image Understanding \textbf{115}(2) (2011)
  224--241

\bibitem{Benenson2015}
Benenson, R., Omran, M., Hosang, J., Schiele, B.
\newblock In: Ten Years of Pedestrian Detection, What Have We Learned? Springer
  International Publishing, Cham (2015)  613--627

\bibitem{Deng2014}
Deng, Y., Luo, P., Loy, C.C., Tang, X.:
\newblock Pedestrian attribute recognition at far distance.
\newblock In: Proceedings of the 22Nd ACM International Conference on
  Multimedia. MM '14, New York, NY, USA, ACM (2014)  789--792

\bibitem{Zhu2016}
Zhu, H., Meng, F., Cai, J., Lu, S.:
\newblock Beyond pixels: A comprehensive survey from bottom-up to semantic
  image segmentation and cosegmentation.
\newblock Journal of Visual Communication and Image Representation \textbf{34}
  (2016)  12--27

\bibitem{Ziebart2009}
Ziebart, B.D., Ratliff, N., Gallagher, G., Mertz, C., Peterson, K., Bagnell,
  J.A., Hebert, M., Dey, A.K., Srinivasa, S.:
\newblock Planning-based prediction for pedestrians.
\newblock In: International Conference on Intelligent Robots and Systems. (Oct
  2009)  3931--3936

\bibitem{VIRATdataset}
Oh, S., Hoogs, A., Perera, A., Cuntoor, N., Chen, C.C., Lee, J.T., Mukherjee,
  S., Aggarwal, J.K., Lee, H., Davis, L., Swears, E., Wang, X., Ji, Q., Reddy,
  K., Shah, M., Vondrick, C., Pirsiavash, H., Ramanan, D., Yuen, J., Torralba,
  A., Song, B., Fong, A., Roy-Chowdhury, A., Desai, M.:
\newblock A large-scale benchmark dataset for event recognition in surveillance
  video.
\newblock In: Computer Vision and Pattern Recognition. (June 2011)  3153--3160

\bibitem{Munoz2010}
Munoz, D., Bagnell, J.A., Hebert, M.:
\newblock Stacked hierarchical labeling.
\newblock In: European Conference on Computer Vision. (2010)  57--70

\bibitem{Yang2014}
Yang, J., Price, B., Cohen, S., Yang, M.H.:
\newblock Context driven scene parsing with attention to rare classes.
\newblock In: Computer Vision and Pattern Recognition. (2014)  3294--3301

\bibitem{Long2015}
Long, J., Shelhamer, E., Darrell, T.:
\newblock Fully convolutional networks for semantic segmentation.
\newblock In: Computer Vision and Pattern Recognition. (2015)  3431--3440

\bibitem{Shelhamer2017}
Shelhamer, E., Long, J., Darrell, T.:
\newblock Fully convolutional networks for semantic segmentation.
\newblock IEEE Transactions on Pattern Analysis and Machine Intelligence
  \textbf{39}(4) (April 2017)  640--651

\bibitem{Huang2016}
Huang, S., Li, X., Zhang, Z., He, Z., Wu, F., Liu, W., Tang, J., Zhuang, Y.:
\newblock Deep learning driven visual path prediction from a single image.
\newblock IEEE Transactions on Image Processing \textbf{25}(12) (Dec 2016)
  5892--5904

\bibitem{Krizhevsky2012}
Krizhevsky, A., Sutskever, I., Hinton, G.E.:
\newblock Imagenet classification with deep convolutional neural networks.
\newblock In Pereira, F., Burges, C.J.C., Bottou, L., Weinberger, K.Q., eds.:
  Advances in Neural Information Processing Systems. (2012)  1097--1105

\bibitem{Bromley1994}
Bromley, J., Guyon, I., LeCun, Y., S{\"a}ckinger, E., Shah, R.:
\newblock Signature verification using a" siamese" time delay neural network.
\newblock In: Advances in Neural Information Processing Systems. (1994)
  737--744

\bibitem{Dalal2005}
Dalal, N., Triggs, B.:
\newblock Histograms of oriented gradients for human detection.
\newblock In: Computer Vision and Pattern Recognition. (2005)  886--893

\bibitem{Enzweiler2010}
Enzweiler, M., Gavrila, D.M.:
\newblock Integrated pedestrian classification and orientation estimation.
\newblock In: Computer Vision and Pattern Recognition. (2010)  982--989

\bibitem{Wei2017}
Ma, W., Huang, D., Lee, N., Kitani, K.M.:
\newblock Forecasting interactive dynamics of pedestrians with fictitious play.
\newblock In: Computer Vision and Pattern Recognition. (2016)  774--782

\bibitem{Singh2012}
Singh, S., Gupta, A., Efros, A.A.:
\newblock Unsupervised discovery of mid-level discriminative patches.
\newblock In: European Conference on Computer Vision. (2012)  73--86

\bibitem{Kitani2012}
Kitani, K.M., Ziebart, B.D., Bagnell, J.A., Hebert, M.:
\newblock Activity forecasting.
\newblock In: European Conference on Computer Vision. (2012)  201--214

\bibitem{Ballan2016}
Ballan, L., Castaldo, F., Alahi, A., Palmieri, F., Savarese, S.:
\newblock Knowledge transfer for scene-specific motion prediction.
\newblock In: European Conference on Computer Vision. (2016)  697--713

\bibitem{Rehder2017}
Rehder, E., Wirth, F., Lauer, M., Stiller, C.:
\newblock Pedestrian prediction by planning using deep neural networks (2017)

\bibitem{Lee2017}
Lee, N., Choi, W., Vernaza, P., Choy, C.B., Torr, P.H.S., Chandraker, M.K.:
\newblock {DESIRE:} distant future prediction in dynamic scenes with
  interacting agents.
\newblock In: Computer Vision and Pattern Recognition. (2017)  336--345

\bibitem{Walker2016}
Walker, J., Gupta, A., Hebert, M.:
\newblock Patch to the future: Unsupervised visual prediction.
\newblock In: Computer Vision and Pattern Recognition. (June 2014)  3302--3309

\bibitem{Park2016}
Park, H.S., Hwang, J.J., Niu, Y., Shi, J.:
\newblock Egocentric future localization.
\newblock In: Computer Vision and Pattern Recognition. (June 2016)  4697--4705

\bibitem{Su2017}
Su, S., Hong, J.P., Shi, J., Park, H.S.:
\newblock Predicting behaviors of basketball players from first person videos.
\newblock In: Computer Vision and Pattern Recognitionr. (2017)  1502--1510

\bibitem{Kooij2014}
Kooij, J.F.P., Schneider, N., Flohr, F., Gavrila, D.M.:
\newblock Context-based pedestrian path prediction.
\newblock In: European Conference on Computer Vision. (2014)  618--633

\bibitem{Schneider2013}
Schneider, N., Gavrila, D.M.:
\newblock Pedestrian path prediction with recursive bayesian filters: A
  comparative study.
\newblock In: German Conference on Pattern Recognition. (2013)  174--183

\bibitem{Xie2013}
Xie, D., Todorovic, S., Zhu, S.C.:
\newblock Inferring {`Dark Matter'} and {`Dark Energy'} from videos.
\newblock In: International Conference on Computer Vision. (Dec 2013)
  2224--2231

\bibitem{Yi2016}
Yi, S., Li, H., Wang, X.:
\newblock Pedestrian behavior understanding and prediction with deep neural
  networks.
\newblock In: European Conference on Computer Vision. (2016)  263--279

\bibitem{Alahi2016}
Alahi, A., Goel, K., Ramanathan, V., Robicquet, A., Fei-Fei, L., Savarese, S.:
\newblock Social lstm: Human trajectory prediction in crowded spaces.
\newblock In: Computer Vision and Pattern Recognition. (June 2016)  961--971

\bibitem{Fernando2017a}
Fernando, T., Denman, S., Sridharan, S., Fookes, C.:
\newblock Soft + hardwired attention: An {LSTM} framework for human trajectory
  prediction and abnormal event detection (2017)

\bibitem{Fernando2017b}
Fernando, T., Denman, S., McFadyen, A., Sridharan, S., Fookes, C.:
\newblock Tree memory networks for modelling long-term temporal dependencies
  (2017)

\bibitem{Lee2016}
Lee, N., Kitani, K.M.:
\newblock Predicting wide receiver trajectories in american football.
\newblock In: Winter Conference on Applications of Computer Vision. (March
  2016)  1--9

\bibitem{Bokhari2017}
Bokhari, S.Z., Kitani, K.M.:
\newblock Long-term activity forecasting using first-person vision.
\newblock In: Asian Conference on Computer Vision. (2016)  346--360

\bibitem{Rhinehart2017}
Rhinehart, N., Kitani, K.M.:
\newblock First-person activity forecasging with online inverse reinforcement
  learning (2017)

\bibitem{Keller2014}
Keller, C.G., Gavrila, D.M.:
\newblock Will the pedestrian cross? a study on pedestrian path prediction.
\newblock IEEE Transactions on Intelligent Transportation Systems
  \textbf{15}(2) (April 2014)  494--506

\bibitem{Rehder2015}
Rehder, E., Kloeden, H.:
\newblock Goal-directed pedestrian prediction.
\newblock In: Workshop on International Conference on Computer Vision. (Dec
  2015)  139--147

\bibitem{Yamaguchi2011}
Yamaguchi, K., Berg, A.C., Ortiz, L.E., Berg, T.L.:
\newblock Who are you with and where are you going?
\newblock In: CVPR 2011. (2011)  1345--1352

\bibitem{Robicquet2016}
Robicquet, A., Sadeghian, A., Alahi, A., Savarese, S.:
\newblock Learning social etiquette: Human trajectory understanding in crowded
  scenes.
\newblock In Leibe, B., Matas, J., Sebe, N., Welling, M., eds.: European
  Conference on Computer Vision, Cham, Springer International Publishing (2016)
   549--565

\bibitem{Helbing1995}
Helbing, D., Molnar, P.:
\newblock Social force model for pedestrian dynamics.
\newblock Physical review E \textbf{51}(5) (1995)  4282

\bibitem{UCYdataset}
Lerner, A., Chrysanthou, Y., Lischinski, D.:
\newblock Crowds by example.
\newblock Computer Graphics Forum \textbf{26}(3) (2007)  655--664

\bibitem{ETHdataset}
Pellegrini, S., Ess, A., Schindler, K., van Gool, L.:
\newblock You'll never walk alone: Modeling social behavior for multi-target
  tracking.
\newblock In: International Conference on Computer Vision. (2009)  261--268

\bibitem{EdinburghDataset}
Majecka, B.:
\newblock Statistical models of pedestrian behaviour in the forum.
\newblock PhD thesis, MSc Dissertation, School of Informatics, University of
  Edinburgh (2009)

\bibitem{TownCentreDataset}
Benfold, B., Reid, I.:
\newblock Stable multi-target tracking in real-time surveillance video.
\newblock In: Computer Vision and Pattern Recognition. (2011)  3457--3464

\bibitem{GrandCentral}
Yi, S., Li, H., Wang, X.:
\newblock Understanding pedestrian behaviors from stationary crowd groups.
\newblock In: Computer Vision and Pattern Recognition. (2015)  3488--3496

\bibitem{KITTI}
Geiger, A., Lenz, P., Urtasun, R.:
\newblock Are we ready for autonomous driving? the kitti vision benchmark
  suite.
\newblock In: Computer Vision and Pattern Recognition. (2012)  3354--3361

\end{thebibliography}

\end{document}